\documentclass[sigconf]{acmart}

\AtBeginDocument{%
  \providecommand\BibTeX{{%
    \normalfont B\kern-0.5em{\scshape i\kern-0.25em b}\kern-0.8em\TeX}}}

\setcopyright{acmcopyright}
\copyrightyear{2021}
\acmYear{2021}
\acmDOI{10.1145/1122445.1122456}

\acmConference[MMSports 2021]{MMSports 2021}{October 20--24, 2021}{Chengdu, China}
\acmBooktitle{MMSports 2021: 4th International ACM Workshop on Multimedia Content Analysis in Sports,
  October 20--24, 2021, Chengdu, China}
\acmPrice{15.00}
\acmISBN{978-1-4503-XXXX-X/18/06}

\usepackage{booktabs}
\usepackage{tabularx}
\newcommand{\bm}[1]{\mathbf{#1}} %Bold vectors and matrices
\usepackage{multirow}
\newcommand\T{{\mathpalette\raiseT\intercal}}
\newcommand\raiseT[2]{%
\setbox0\hbox{$#1{#2}$}\raise\dp0\box0}

\usepackage{algorithm,algorithmic}
\makeatletter
\newcommand{\ALOOP}[1]{\ALC@it\algorithmicloop\ #1%
  \begin{ALC@loop}}
\newcommand{\ENDALOOP}{\end{ALC@loop}\ALC@it\algorithmicendloop}

\makeatother
\begin{document}

\title{STAR: Noisy Semi-Supervised Transfer Learning for Visual Classification}

%\email{webmaster@marysville-ohio.com}
%\affiliation{%
%  \institution{Institute for Clarity in Documentation}
%  \streetaddress{P.O. Box 1212}
%  \country{USA}

\author{Hasib Zunair}
\affiliation{%
  \institution{Concordia University}
  %\streetaddress{1 Th{\o}rv{\"a}ld Circle}
  \city{Montreal}
  \country{Canada}}

\author{Yan Gobeil}
\affiliation{%
  \institution{D\'ecathlon}
  \city{Montreal}
  \country{Canada}
}

\author{Samuel Mercier}
\affiliation{%
 \institution{D\'ecathlon}
 %\streetaddress{Rono-Hills}
 \city{Montreal}
 %\state{Arunachal Pradesh}
 \country{Canada}}

\author{A. Ben Hamza}
\affiliation{%
  \institution{Concordia University}
  %\streetaddress{30 Shuangqing Rd}
  \city{Montreal}
  %\state{Beijing Shi}
  \country{Canada}}
 \email{hamza@ciise.concordia.ca}

%\affiliation{%
%  \institution{Palmer Research Laboratories}
%  \streetaddress{8600 Datapoint Drive}
%  \country{USA}
%\affiliation{%
%  \institution{The Th{\o}rv{\"a}ld Group}
%  \streetaddress{1 Th{\o}rv{\"a}ld Circle}
%\email{jsmith@affiliation.org}
%\affiliation{%
%  \institution{The Kumquat Consortium}
%  \city{New York}
%\email{jpkumquat@consortium.net}


\renewcommand{\shortauthors}{}

\begin{abstract}
Semi-supervised learning (SSL) has proven to be effective at leveraging large-scale unlabeled data to mitigate the dependency on labeled data in order to learn better models for visual recognition and classification tasks. However, recent SSL methods rely on unlabeled image data at a scale of billions to work well. This becomes infeasible for tasks with relatively fewer unlabeled data in terms of runtime, memory and data acquisition. To address this issue, we propose noisy semi-supervised transfer learning, an efficient SSL approach that integrates transfer learning and self-training with noisy student into a single framework, which is tailored for tasks that can leverage unlabeled image data on a scale of thousands. We evaluate our method on both binary and multi-class classification tasks, where the objective is to identify whether an image displays people practicing sports or the type of sport, as well as to identify the pose from a pool of popular yoga poses. Extensive experiments and ablation studies demonstrate that by leveraging unlabeled data, our proposed framework significantly improves visual classification, especially in multi-class classification settings compared to state-of-the-art methods. Moreover, incorporating transfer learning not only improves classification performance, but also requires 6x less compute time and 5x less memory. We also show that our method boosts robustness of visual classification models, even without specifically optimizing for adversarial robustness.
\end{abstract}

\begin{CCSXML}
<ccs2012>
 <concept>
  <concept_id>10010520.10010553.10010562</concept_id>
  <concept_desc>Computer systems organization~Embedded systems</concept_desc>
  <concept_significance>500</concept_significance>
 </concept>
 <concept>
  <concept_id>10010520.10010575.10010755</concept_id>
  <concept_desc>Computer systems organization~Redundancy</concept_desc>
  <concept_significance>300</concept_significance>
 </concept>
 <concept>
  <concept_id>10010520.10010553.10010554</concept_id>
  <concept_desc>Computer systems organization~Robotics</concept_desc>
  <concept_significance>100</concept_significance>
 </concept>
 <concept>
  <concept_id>10003033.10003083.10003095</concept_id>
  <concept_desc>Networks~Network reliability</concept_desc>
  <concept_significance>100</concept_significance>
 </concept>
</ccs2012>
\end{CCSXML}

\ccsdesc{Computing methodologies~Artificial intelligence~Computer vision}

\keywords{Semi-supervised learning, self-training, transfer learning, visual classification}


\maketitle

\section{Introduction}
Visual classification is a fundamental problem in computer vision. It refers to the process of organizing a dataset of images into a known number of classes, and the task is to assign new images to one of these classes. Conventional machine learning models for visual classification rely on labeled data to train a classifier~\cite{krizhevsky2012imagenet}. While unlabeled data can be generally obtained with minimal human labor, labeling data is often laborious, costly and requires the efforts of experienced human annotators. Semi-supervised learning (SSL) addresses this issue by leveraging large volumes of unlabeled data together with a relatively small amount of labeled data in a bid to learn better visual classifiers~\cite{zhu2005semi}. SSL has had a resurgence in recent years, in large part thanks to its ability to improve the accuracy of the model on important benchmarks. The objective of semi-supervised learning for the classification of images is to predict the most probable labels of images in a dataset by leveraging labeled and unlabeled data in order to train a classifier.

Some of the early approaches to semi-supervised visual classification have focused mostly on low-level computer vision frameworks, which involve hand-engineered features. These methods include co-training~\cite{blum1998combining} and self-training~\cite{scudder1965probability} approaches, as well as transductive support vector machines and graph-based techniques~\cite{liu2010large}. Graph-based methods, for instance, capture the manifold structure of the data and encourage similar points to share labels. A major limitation with graph-based approaches is the need for similarity measures that create graphs with no inter-class connections. In many real-world applications, it is not easy to learn such a good visual similarity metric. Moreover, intra-class variations are often larger than inter-class variations, making pairwise similarity based methods of little utility. In addition, hand-crafted features often lead to unsatisfactory results on unseen data and do not generalize well across different tasks.

The advent of deep learning has sparked groundswell of interest in the adoption of deep neural networks (DNNs) for semi-supervised visual classification. A plethora of DNNs is based on convolutional neural networks, including pre-trained models such as InceptionV3, ResNet, and EfficientNet~\cite{szegedy2016rethinking,he2016deep,tan2019efficientnet,szegedy2015going}, as well
as SSL models with large convolutional networks based on a teacher/student paradigm by training on the labeled data to get an initial teacher model, followed by training a student model~\cite{yalniz2019billion}. These pre-trained models can be used for visual classification, feature extraction, and fine-tuning. Using a pre-trained network with transfer learning is typically much faster and easier than training a network from scratch.

In this paper, we introduce noisy \underline{s}emi-supervised \underline{t}r\underline{a}nsfer lea\underline{r}ning (STAR), an efficient SSL framework for visual classification with the goal of remedying the aforementioned  issues. Our approach mitigates the issue of time required in training large models, while improving classification performance and reducing overfitting. The proposed approach is well suited for tasks that can leverage unlabeled image data on a scale of thousands, and is comprised of two main integrated stages. In the first stage, we train a supervised learning model, pre-trained on the ImageNet database~\cite{deng2009imagenet}, on the labeled data. This model is then used to generate pseudo-labels for the unlabeled data. In the second stage, a larger pre-trained model is trained on the combined labeled data and pseudo-labeled data. The larger model is noised using data augmentation and dropout~\cite{srivastava2014dropout} during training. The main contributions of this paper can be summarized as follows:
\begin{itemize}
\item We propose a noisy semi-supervised transfer learning framework by integrating transfer learning and noisy student training with the goal of improving visual classification performance and reducing computation overhead.
\item We show through extensive experiments that our approach yields significant improvements over strong baseline methods on binary and multi-class classification tasks. These improvements are not only in terms of classification performance, but also in terms of computation time and memory required to train models for the desired tasks.
\item We show that our method boosts robustness in visual classification models without specifically optimizing for adversarial robustness.

\end{itemize}
The rest of this paper is organized as follows. In Section 2, we review important relevant work. In Section 3, we present our STAR framework, which couples transfer learning and noisy student training to jointly improve classification performance and reduce computation overhead. In Section 4, we present experimental results to demonstrate the competitive performance of our approach on both binary and multi-class classification tasks. Finally, we conclude in Section 5 and point out future work directions.

\section{Related Work}
Semi-supervised learning refers to the task of learning a prediction rule from a small amount of labeled data and a large amount of unlabeled data in order to improve model performance. It aims at bridging the gap between unsupervised learning, which uses unlabeled training data, and supervised learning, which uses labeled training data. Convolutional neural networks (CNNs) have become the de facto model for semi-supervised learning in image classification tasks. Sohn \textit{et al}~\cite{sohn2020fixmatch} introduce a semi-supervised learning method that combines pseudo-labeling and consistency regularization. Pseudo-labeling effectively uses the model's class prediction as a label to train against, while consistency regularization assumes that a model should output similar predictions when fed perturbed versions of the same image. Xie \textit{et al}~\cite{xie2020self} present self-training with noisy student or noisy student training, which extends the idea of self-training and distillation with the use of equal-or-larger student models and noise added to the student during learning which has achieved state-of-the-art results on ImageNet~\cite{deng2009imagenet}. Yalniz \textit{et al}~\cite{yalniz2019billion} proposed a learning method based on the teacher/student paradigm which leverages large collection of unlabeled images. Their method is not only improves standard architectures for image classification, but also improves performance for video classification and fine-grain recognition. SSL methods based on generative adversarial networks (GANs) have also received considerable attention~\cite{dai2017good,li2019semi}. Dai \textit{et al}~\cite{dai2017good} propose a semi-supervised learning method based on GANs. The method uses generated data to improve performance for the desired task. A key finding in their study is that a bad generator improves generalization. Li \textit{et al}~\cite{li2019semi} consider GAN-based semi-supervised learning by comparing methods such as Triple GAN and Bad GAN. A generative network is trained on a set of data to produce or replicate similar examples using a generator and a discriminator. The goal of these methods is to predict the labels for unlabeled data, while generating new samples conditioned on these labels. They find that both methods can be used for different purposes. Triple GAN generates good image and label pairs, which can be used to train a classifier, whereas Bad GAN generates samples that force a shift in the decision boundary between the data manifold of the different classes.

Although this burgeoning literature has provided many useful insights, several gaps remain between model architecture design and training on image datasets at different scales (e.g. samples in thousands or billions). Recent state-of-the-art methods rely on unlabeled data at a scale of billions to work well. This becomes infeasible for tasks that can leverage unlabeled image data on a scale of thousands in terms of runtime, memory and data acquisition. In our work, we target the limitations of existing deep SSL approaches and aim to develop an integrated framework for visual classification by leveraging unlabeled data on a scale of thousands in order to learn better classification models.

\section{Method} \label{Method}
\noindent\textbf{Problem Statement.}\quad Given the labels of a small subset of an image dataset, the objective of semi-supervised learning is to leverage unlabeled image data to improve model performance. More specifically, let $\mathcal{D}_{\ell}=\{(\bm{x}_i,y_i)\}_{i=1}^{N_{\ell}}$ be the set of labeled images $\bm{x}_{i}$ with associated known labels $y_i\in\mathcal{Y}_\ell$, and $\mathcal{D}_{u}=\{\bm{x}_i\}_{i=N_{\ell}+1}^{N_{\ell}+N_{u}}$ be the set of unlabeled image data, where $N_{\ell}+N_{u}=N$. Then, the problem of semi-supervised image classification is to learn a classifier with the goal of predicting the labels of the set $\mathcal{D}_{u}$. It is important to note that for multi-class classification problems, the label of each image $\bm{x}_i$ in the labeled set $\mathcal{D}_{\ell}$ can be represented as a $C$-dimensional one-hot vector $\bm{y}_{i} \in \{0, 1\}^{C}$, where $C$ is the number of classes.

\begin{figure*}[!htb]
\centering
\begin{tabular}{cc}
\includegraphics[scale=.8]{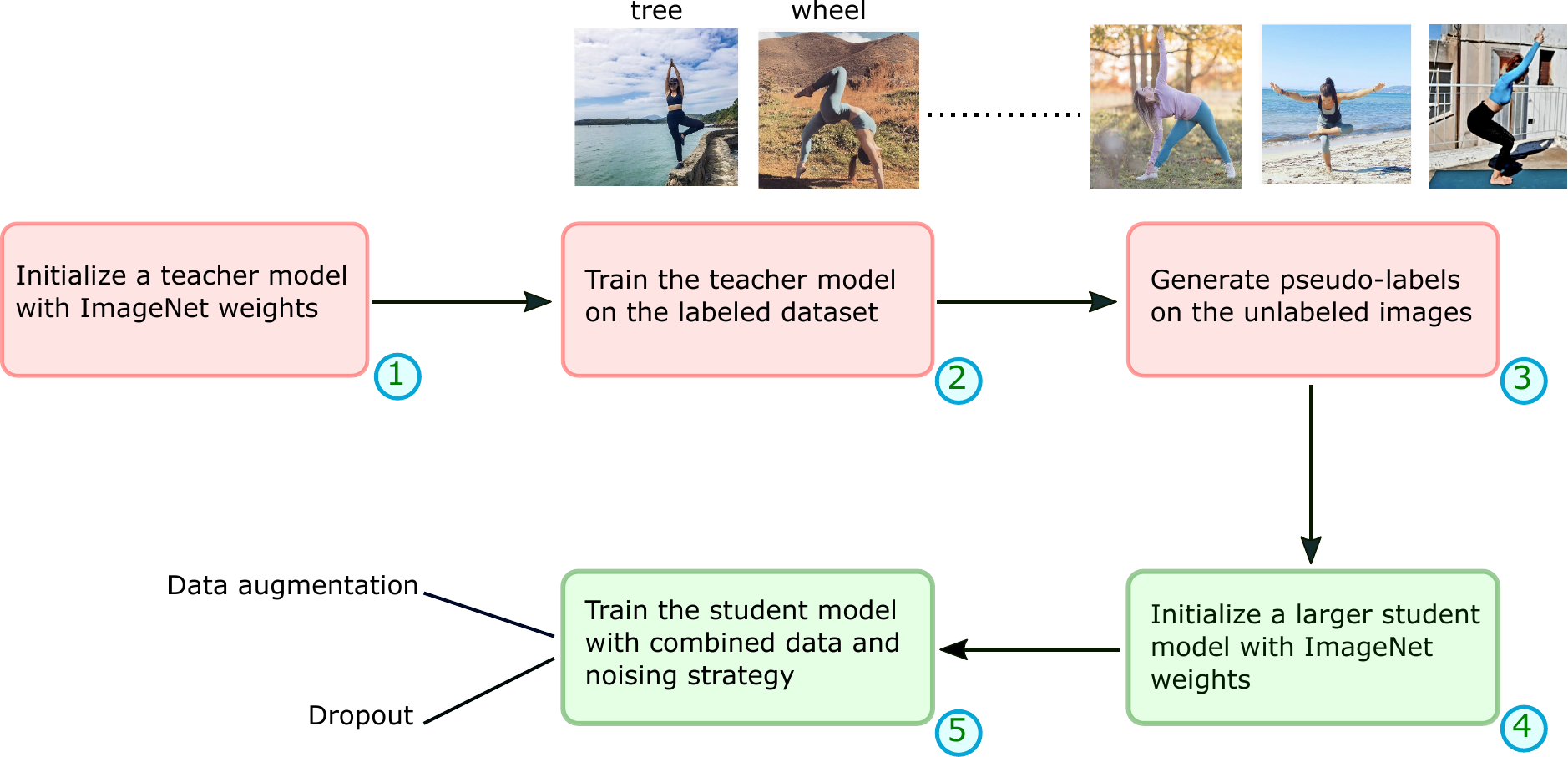}
\end{tabular}
\caption{Schematic layout of the proposed noisy semi-supervised transfer learning (STAR) framework.}
\label{fig:STAR}
\end{figure*}

\medskip\noindent We now present the main building blocks of our noisy semi-supervised transfer learning framework, which integrates transfer learning and noisy student training, as illustrated in Figure~\ref{fig:STAR}. Transfer learning is tailored for tasks that can leverage unlabeled data of a scale of thousands for visual classification, while noisy student training is designed for training models at a very large scale (e.g. on ImageNet, noisy student training uses a teacher model to generate pseudo-labels for 300M unlabeled images). Our proposed STAR method first enables the teacher model to learn more effective representations with the knowledge from prior pre-training on ImageNet via transfer learning. This makes the student model learn even better representations than the teacher model by taking advantage of transfer learning, noise and the unlabeled data. The schematic layout and main steps of the proposed framework are illustrated in Figure~\ref{fig:STAR}.

\medskip\noindent\textbf{Transfer Learning.}\quad Due to limited training data, it is  standard  practice to leverage deep learning models that were pre-trained on large datasets~\cite{yosinski2014transferable}. In our experiments, we use EfficientNets\cite{tan2019efficientnet} pre-trained on ImageNet~\cite{deng2009imagenet}. EfficientNets are a family of pre-trained convolutional neural networks that use a compound scaling method to uniformly scale the network width, depth, and resolution. We replace the final fully connected (FC) layer of the pre-trained model with a global average pooling (GAP) layer, which is widely used in classification tasks. It computes the average output of each feature map in the previous layer and helps minimize overfitting by reducing the total number of parameters in the model. GAP turns a feature map into a single number by taking the average of the numbers in that feature map. Similar to max pooling layers, GAP layers have no trainable parameters and are used to reduce the spatial dimensions of a three-dimensional tensor. The GAP layer is followed by a hidden layer and a single FC layer with a softmax function (i.e. a dense softmax layer of multiple units for the binary or multi-class classification case) that yields the probabilities of predicted classes. We train the models in two stages. First, we replace and train the newly added layers. Then, we also fine-tune the weights of the pretrained model by continuing the backpropagation. In the second stage, we only fine-tune the last two blocks of the pretrained model.

\medskip\noindent\textbf{Noising Student.}\quad We use data augmentation and dropout as a form of input noise when training the student models. This strategy is usually carried out to  improve generalization performance in classification tasks~\cite{zunair2020melanoma,frid2018synthetic}. It is often done by creating modified versions of the input images in a dataset through random transformations, including horizontal and vertical flip, Gaussian noise, brightness and zoom augmentation, horizontal and vertical shift, sampling noise once per pixel, color space conversion, and rotation. For model noise, we use dropout~\cite{srivastava2014dropout}, which can be viewed as a way of regularizing a deep neural network by adding noise to its hidden units. Applying input noise and model noise to the unlabeled data enables the student model to treat the labeled and unlabeled data as a single distribution. Also, data augmentation enforces the student model to make consistent predictions across augmented versions of an image.  While the teacher model generates pseudo-labels for clean images, the student model is trained to predict those labels for an augmented image as input. In other words, the student model is forced to retain the same category for the augmented image as the original image. When using dropout, the layer drops the neuron connection with a certain probability. This makes the teacher model like an ensemble of models when it generates pseudo-labels. We find that the effects of noising the student corroborate with the findings reported in~\cite{xie2020self}.

\medskip\noindent\textbf{Algorithm.}\quad The main algorithmic steps of our approach are summarized in Algorithm~\ref{algo:algoooo}. The input is a set of labeled and unlabeled images for a particular visual classification task. The goal is to use the labeled and unlabeled data for the underlying task. A supervised learning model (i.e. teacher model), which has been pre-trained on ImageNet, is trained on the labeled images by minimizing the cross-entropy loss. Then, the teacher model is used to generate hard pseudo-labels (one-hot encodings) for the unlabeled data. Using both the labeled and pseudo-labeled data, a larger pre-trained model (e.g. student model) is trained to minimize the cross-entropy loss. During this training process, data augmentation and dropout are used as a form of input noise and model noise.

\begin{algorithm}
\caption{Noisy semi-supervised transfer learning (STAR)}
\label{algo:algoooo}
\begin{algorithmic}[1]
\REQUIRE Labeled set $\mathcal{D}_{\ell}=\{(\bm{x}_i,y_i)\}_{i=1}^{N_{\ell}}$ of images $\bm{x}_{i}$ with associated known labels $y_i$, and unlabeled set $\mathcal{D}_{u}=\{\bm{x}_i\}_{i=N_{\ell}+1}^{N_{\ell}+N_{u}}$ for a task $\mathcal{{T}}$.
\ENSURE Learned parameters of the student model for task $\mathcal{{T}}$.
\STATE Initialize a supervised teacher model $f_t$ with ImageNet weights.
\STATE Train $f_t$ on the labeled image set by minimizing the cross-entropy loss.
\STATE Use $f_t$ to generate pseudo-labels for the unlabeled image set.
\STATE Initialize a \textbf{larger} supervised student model $f_s$ with ImageNet weights.
\STATE Train $f_s$ with added \textbf{noise} on the combined labeled and pseudo-labeled images by minimizing the cross-entropy loss.
\end{algorithmic}
\end{algorithm}

\section{Experiments} \label{Experiments}
In this section, we conduct extensive experiments to assess the performance of the proposed visual classification framework in comparison with state-of-the-art methods on several datasets. The source code to reproduce the experimental results is made publicly available on GitHub\footnote{https://github.com/Decathlon/decavision}.

\subsection{Experimental Setup}
\noindent\textbf{Datasets.}\quad The summary descriptions of the datasets used to demonstrate and analyze the performance of the proposed STAR method are as follows:
\begin{itemize}
\item \textbf{Labeled datasets:} We conduct experiments on the Sport-or-not, Yoga-Pose, and Sport datasets.  The Sport-or-not dataset is comprised of two classes with around 14K images being either of people practicing or not practicing sport. The Yoga-Pose dataset consists of almost 3K images of 18 different popular yoga poses such as \textit{bridge}, \textit{lotus}, and \textit{tree}. The Sport dataset consists of 155 classes of popular sports such as \textit{axe-throwing}, \textit{basketball}, \textit{cricket}, \textit{kayaking} and many more. The labeled datasets are split into training, validation and test sets. For the Yoga-Pose dataset, the test set consists only of images that are in general harder to classify. A separate set is used as an additional test set, termed Test 2, only for the Sport-or-not dataset, as the classification task is much easier and less ambiguous. All datasets consist of training sets that are balanced (except for the Sport dataset) and test sets that are almost class balanced. Dataset statistics are summarized in Table~\ref{Tab:datastats}.

\item \textbf{Unlabeled datasets:} We obtain two sets of unlabeled data termed Unlb-700 and Unlb-120 from different sources, consisting of over 700K and 120K images, respectively. While the labeled data are collected from Instagram and Google Images, the unlabeled data are collected from Instagram using hashtag \textit{decathlon} for sport images and various hashtags for yoga poses, and Getty Images for yoga poses using appropriate search keywords. Although these images are acquired using search keywords or labels, we ignore the labels and treat them as unlabeled data. For Sport-or-not and Sport classification tasks, we use the Unlb-700 dataset. For Yoga-Pose classification task, we use the Unlb-120 dataset.
\end{itemize}

\begin{table}[!htb]
	\caption{Dataset statistics for Sport-or-not, Yoga-Pose and Sport classification.}
	\label{Tab:datastats}
	\setlength{\tabcolsep}{3.8pt}
	\centering
	\medskip
	\begin{tabular}{@{}lrrrrr@{}}
		\toprule
		& \multicolumn{5}{c}{Number of samples} \\
		\cmidrule(lr){3-6}
		Dataset & Classes & Train & Validation & Test 1 & Test 2\\
		\midrule
		Sport-or-not ~ & 2 & 12,366 & 1,002 & 1,002 & 1,000 \\
		Yoga-Pose ~ & 18 & 2,855 & 196 & 214 & - \\
		Sport ~ & 155 & 74,961 & 12,830 & 3,591 & - \\
		\bottomrule
		\hline
	\end{tabular}
\end{table}

Prior to training all models, the training and validation sets are converted to the TFRecords format with the desired input size (i.e. $299\times 299\times 3$) through image resizing. In order to achieve faster convergence, feature standardization is usually performed, i.e. we rescale the images to have values between 0 and 1. More specifically, given a data matrix $\bm{X}=(\bm{x}_{1},\ldots,\bm{x}_{n})^{\T}$, the standardized feature vector is given by
\begin{equation}
\bm{z}_{i} = \frac{\bm{x}_{i} - \min(\bm{x}_{i})}{\max(\bm{x}_{i}) - \min(\bm{x}_{i})},\quad i=1,\ldots,n,
\label{eq:std}
\end{equation}
where $\bm{x}_{i}$ is the $i$-th input data point, denoting a row vector. It is important to note that in our approach, no post-processing is employed other than resizing and feature standardization.

\medskip\noindent\textbf{Baselines.}\quad We evaluate the performance of the proposed method against EfficientNets~\cite{tan2019efficientnet} and Noisy Student Training~\cite{xie2020self}. A brief description of these state-of-the-art baselines can be summarized as follows:
\begin{itemize}
\item \textbf{EfficientNets}\quad As the name suggests, EfficientNets is a group of computationally efficient convolutional neural networks that are trained on the ImageNet database, and they use a compound scaling method to uniformly scale the network width, depth, and resolution with a set of fixed scaling coefficients~\cite{tan2019efficientnet}. This group consists of eight models with configurations ranging from the baseline model EfficientNet-B0 to the larger model EfficientNet-B7, where each subsequent architecture refers to a model variant with more parameters and higher accuracy. Depth is the number of layers in the network, width is the number of filters in a convolutional layer, and resolution is simply the height and width of the input image. The main building block of EfficientNets is the mobile inverted bottleneck convolution (MBConv) together with a squeeze-and-excitation optimization.

\item \textbf{Noisy Student Training (NST)}\quad is a semi-supervised learning method for visual classification~\cite{xie2020self}. The method consists of two main stages. First, a supervised model is trained from scratch on the labeled images, and then then used as a teacher model to generate the pseudo-labels for the unlabeled data. Second, a larger model, known as student, is trained from scratch on the combined of labeled and pseudo-labeled image data. Noise is injected into the student model in the form of dropout and data augmentation with the goal of achieving better generalization than the teacher model. This process is then repeated until a desired level of classification performance is achieved. Unlike NST in which the student becomes the teacher durring the iterative training process, our STAR approach uses pre-training for both teacher and student models.
\end{itemize}

\medskip\noindent\textbf{Implementation Details.}\quad All experiments are performed using the DecaVision library on Linux workstations running 4.8Hz and 64GB RAM with NVIDIA RTX 2080Ti and RTX 3080 GPUs. All models are based on EfficientNets~\cite{tan2019efficientnet}, particularly EfficientNet-B3, and larger models such as EfficientNet-B5 and EfficientNet-B7 which have more learnable parameters to fit large number of unlabeled data. Pretrained ImageNet~\cite{krizhevsky2012imagenet} models are trained for our tasks using the Adam optimization algorithm~\cite{kingma2014adam} to minimize the binary or categorical cross entropy losses, depending on the binary or multi-class setting. A factor of 0.1 is used to reduce the learning rate once the loss stagnates. Training is continued until the validation loss stagnates using an early stopping mechanism.
For Sport-or-not and Yoga-Pose classification tasks, we use hard pseudo-labels (one-hot encoding). For the Sport classification task, we use all unlabeled data that has a confidence score higher than a threshold equal to 0.3.

Due to the nature of deep learning algorithms, we train every model in two stages. In the first stage, we perform a hyperparameter optimization process using the hypertuning feature in the DecaVision library, which is inspired by the scikit-optimize library\footnote{https://scikit-optimize.github.io/stable/}. We start by training a model 10 times with random combinations of hyperparameters, which are predefined in the search space (i.e. hidden size, learning rate, whether to skip or not the fine-tuning phase, learning rate in the fine-tuning phase, etc.). Then, we use what has been learned from these random combinations to find 15 better ones. In total, a single model configuration goes through 25 iterations of hyperparameter search to find the best model configuration possible. In the second stage, we train a new model configuration using the optimal hyperparameters. This stage starts by training an extra layer or layers, depending on the hyperparameter optimization, on top of the frozen pretrained model, followed by fine-tuning few blocks of the pretrained model by unfreezing them.

\medskip\noindent\textbf{Evaluation metrics.}\quad In order to evaluate the performance of our proposed STAR framework against the baseline methods, we use average accuracy and F1 score as evaluation metrics. For the binary classification task (i.e. Sport-or-not classification), we measure performance using the average accuracy since the validation and test set are balanced, as shown in Table~\ref{Tab:datastats}. For the multi-class classification tasks (i.e. Yoga-Pose and Sport classification), we report both average accuracy and per-class F1 score. For both metrics, a larger value indicates better classification performance.

\subsection{Results and Analysis} \label{Results}
In this subsection, we evaluate the proposed STAR approach on several datasets, demonstrating significant improvements over strong baseline methods.

\begin{figure}[!htb]
\setlength{\tabcolsep}{.1em}
\centering
\begin{tabular}{ccc}
\includegraphics[width=.95in,height=.95in]{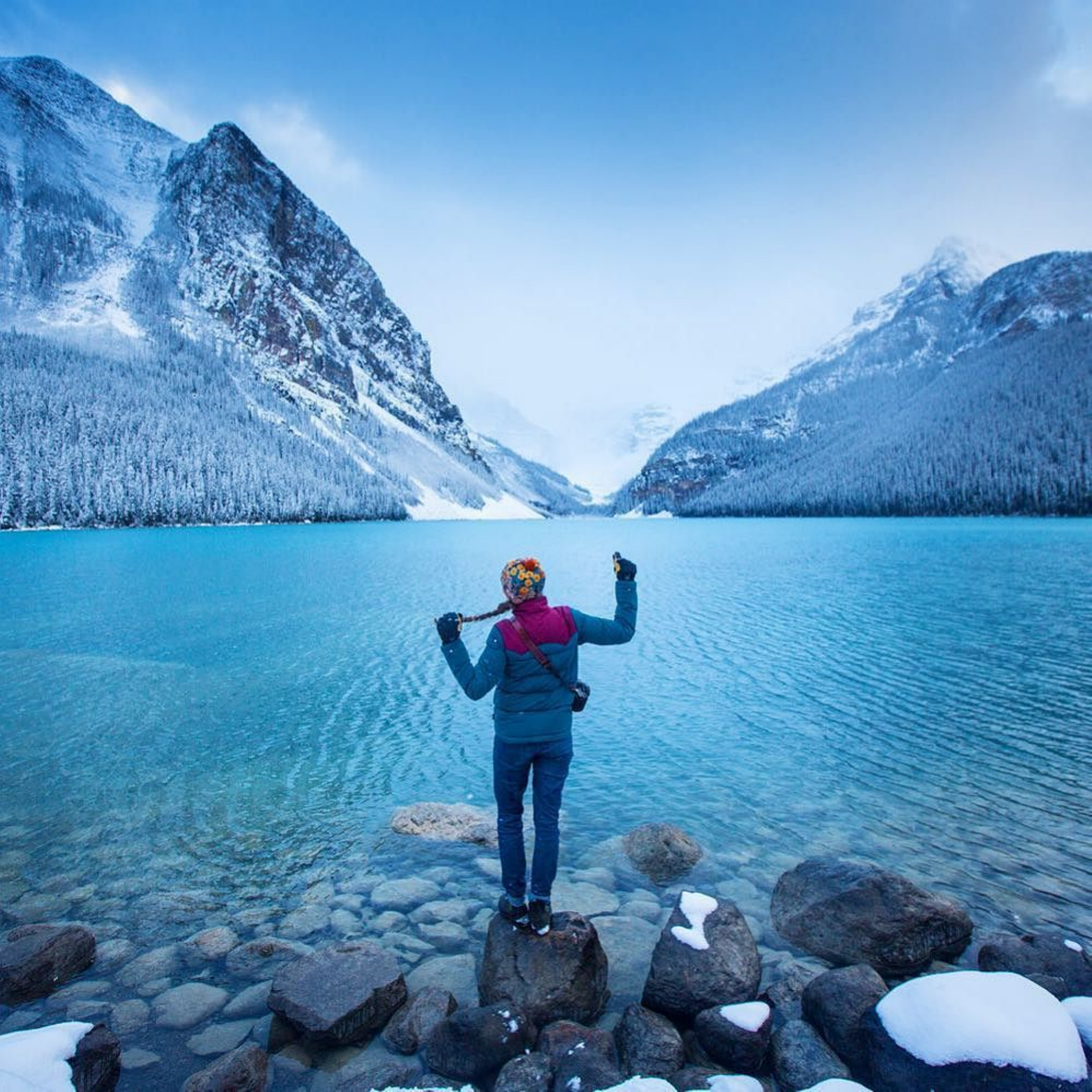} & \includegraphics[width=1.35in,height=.95in]{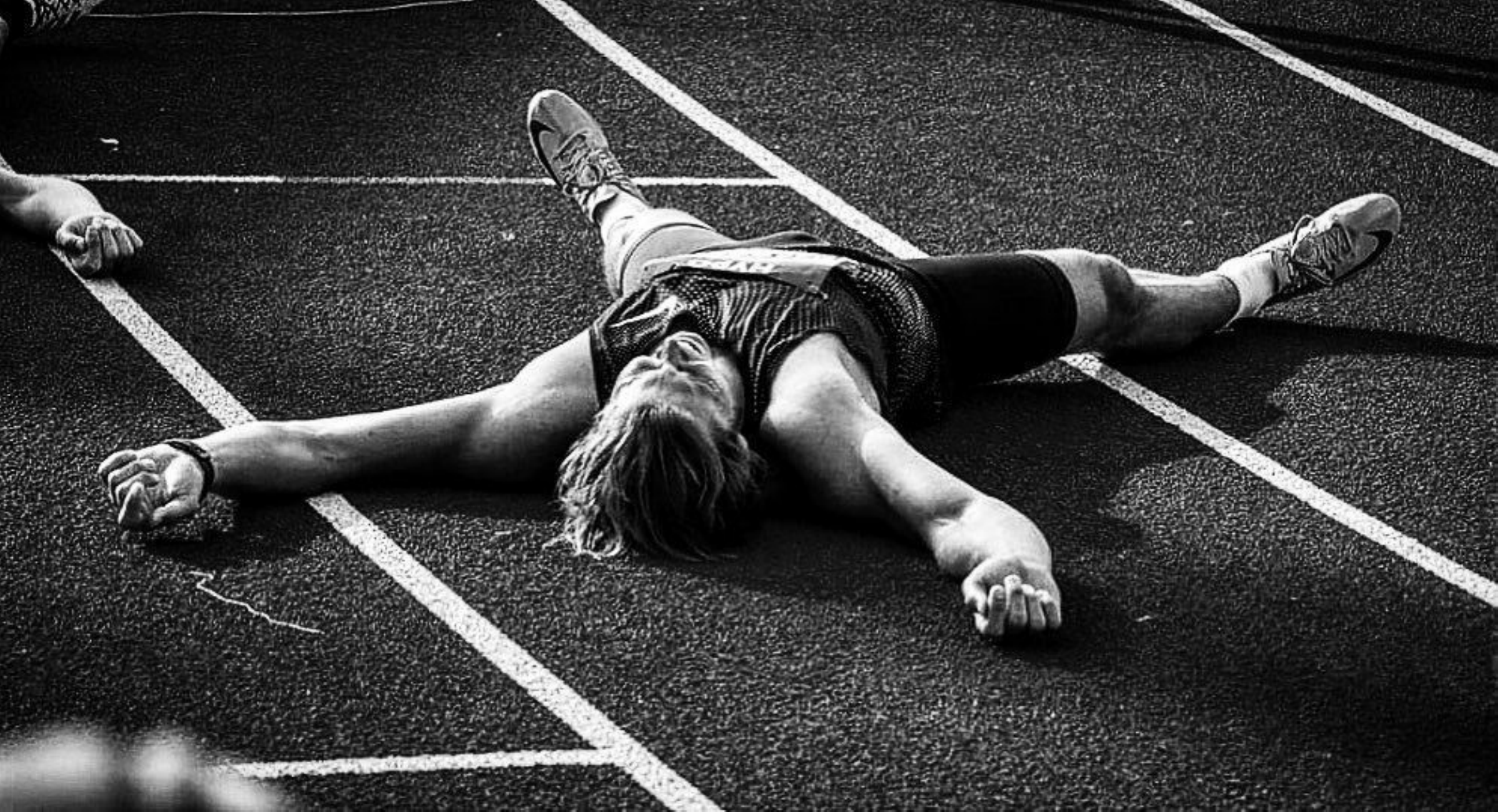} & \includegraphics[width=.95in,height=.95in]{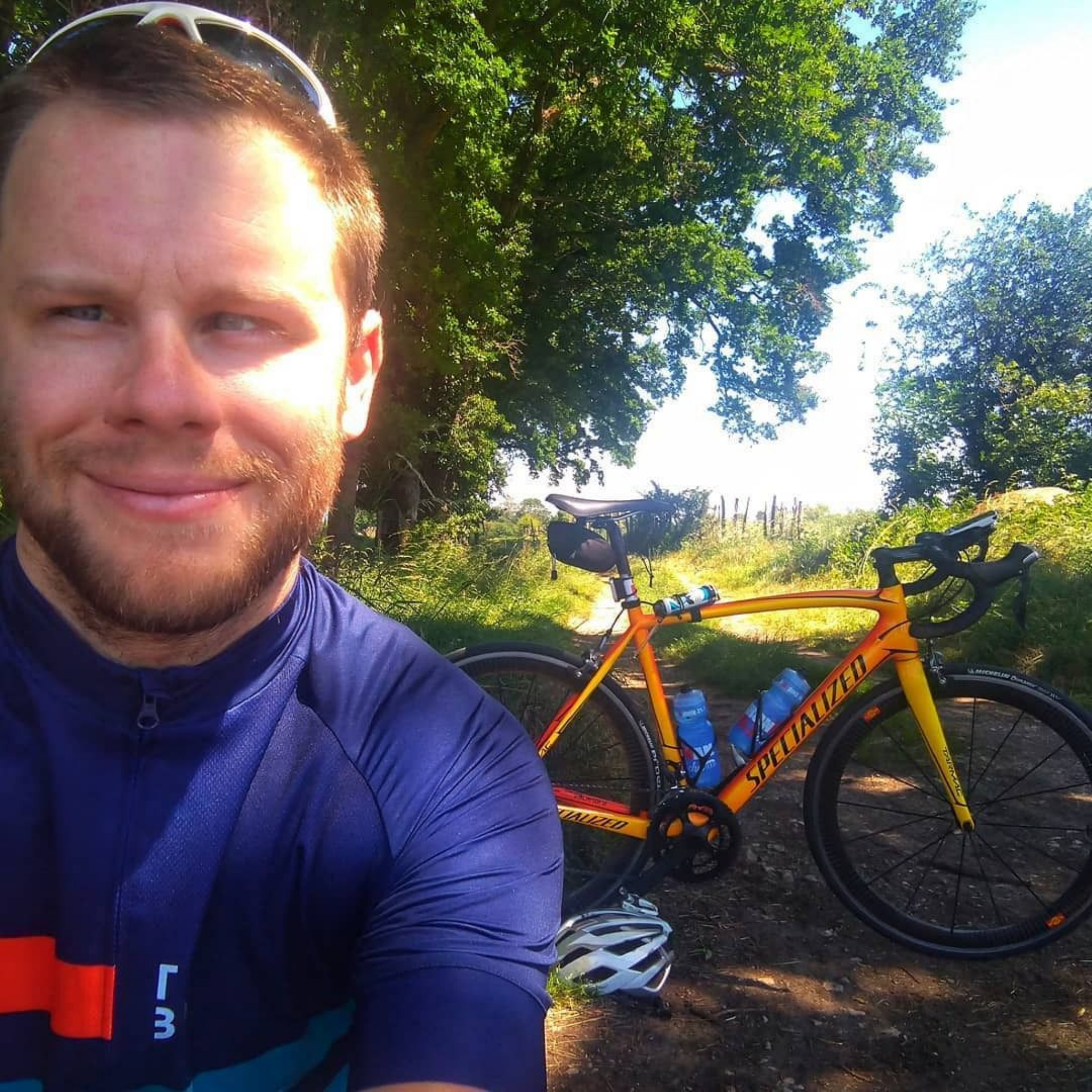}

\end{tabular}
\caption{Example images of people not practicing sports, but have regions representative of sports and vice versa. It is unclear whether to label these images as \textit{sport} or \textit{not-sport} even by a human annotator.}
\label{fig:amb}
\end{figure}

\medskip\noindent\textbf{Sport-or-not classification results} \quad In Table~\ref{Tab:sporornot}, we report the classification results on the Sport-or-not dataset. The teacher model is the EfficientNet-B3 architecture trained on the labeled dataset. Pseudo-labels are then generated on the Unlb-700 dataset. This is followed by training an EfficientNet-B5 architecture on a combination of the labeled and pseudo-labeled datasets with data augmentation as noise. Both models were pre-trained on ImageNet. Notice that accuracy on the test set drops even after training on more than half a million images. This drop in accuracy is largely attributed to the fact that many images in the test set are very ambiguous and it is not easy for a human annotator to label such an image as \textit{sport} or \textit{not-sport}. A typical example is a picture of someone standing in front of the mountains, as shown in Figure~\ref{fig:amb} (left).

Figure~\ref{fig:amb} (left) can be considered as \textit{sport} if the person is mountain-climbing/hiking, or \textit{not-sport} if the person is just taking a picture with mountains in the background. Figure~\ref{fig:amb} (center) shows a person not actually doing sport, but clearly was practicing sport moments earlier. Figure~\ref{fig:amb} (right) shows a person is biking, but not in the image. So the question is: can you tell if these images are \textit{sport} or \textit{not-sport}? This motivated us to build another unseen balanced test set, which we term Test 2, with only unambiguous images in an effort to properly evaluate our models. On the Test 2 set, the student model outperforms the teacher model by a relative improvement of 2.4\% in terms of average accuracy. Also, our STAR approach yields a relative improvement of 1.36\% over NST.

\begin{table}[!htb]
\caption{Classification performance on the Sport-or-not datasets with two classes. For the teacher model, we train EfficientNet-B3 on the labeled training set. Both NST and our STAR method are trained on a combination of pseudo-labeled data, generated by EfficientNet-B3 on the unlabeled Unlb-700 dataset, and the labeled data. Boldface numbers indicate the best performance.}
	\label{Tab:sporornot}
	\setlength{\tabcolsep}{3.8pt}
	\centering
	\medskip
	\begin{tabular}{@{}lcccccccccccc@{}}
		\toprule
		& \multicolumn{3}{c}{Accuracy (\%)} \\
		\cmidrule(lr){2-4}
		Method & Validation & Test 1 & Test 2 \\
		\midrule
		EfficientNet-B3~\cite{tan2019efficientnet} ~ & 86.03 & \textbf{86.58} & 87.20 \\
		EfficientNet-B5 NST~\cite{xie2020self} ~ & 86.21 & 85.36 & 88.10 \\
		EfficientNet-B5 STAR ~ & \textbf{86.53} & 85.88 & \textbf{89.30} \\
		\bottomrule
		\hline
	\end{tabular}
\end{table}

\begin{figure*}[!htb]
\centering
\includegraphics[scale=.4]{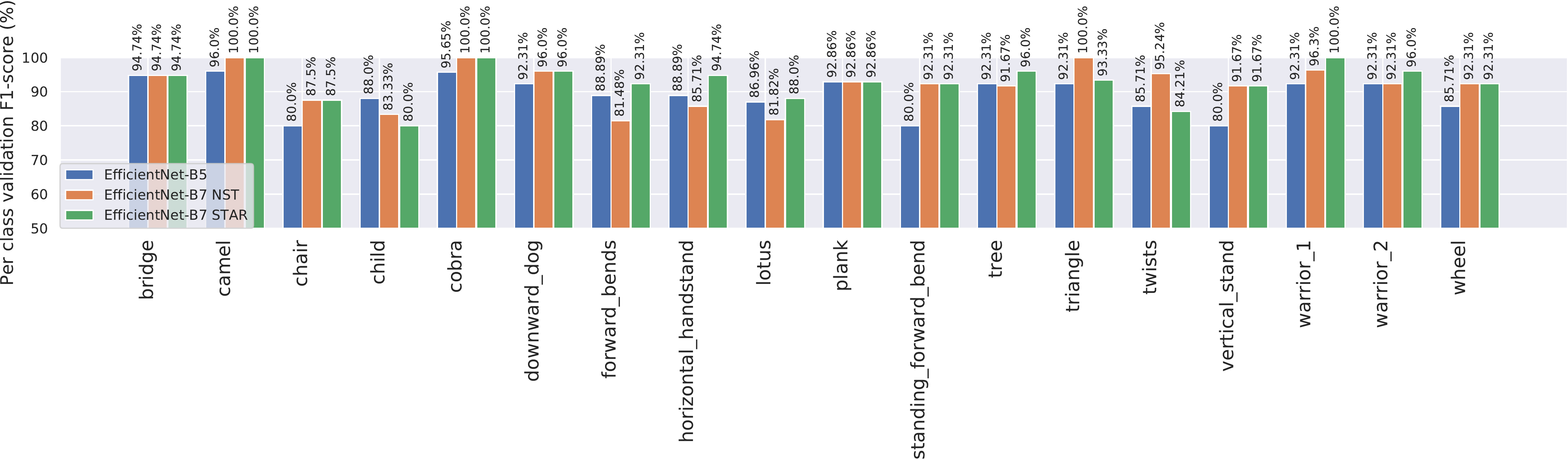}\\[0.1ex]
\includegraphics[scale=.4]{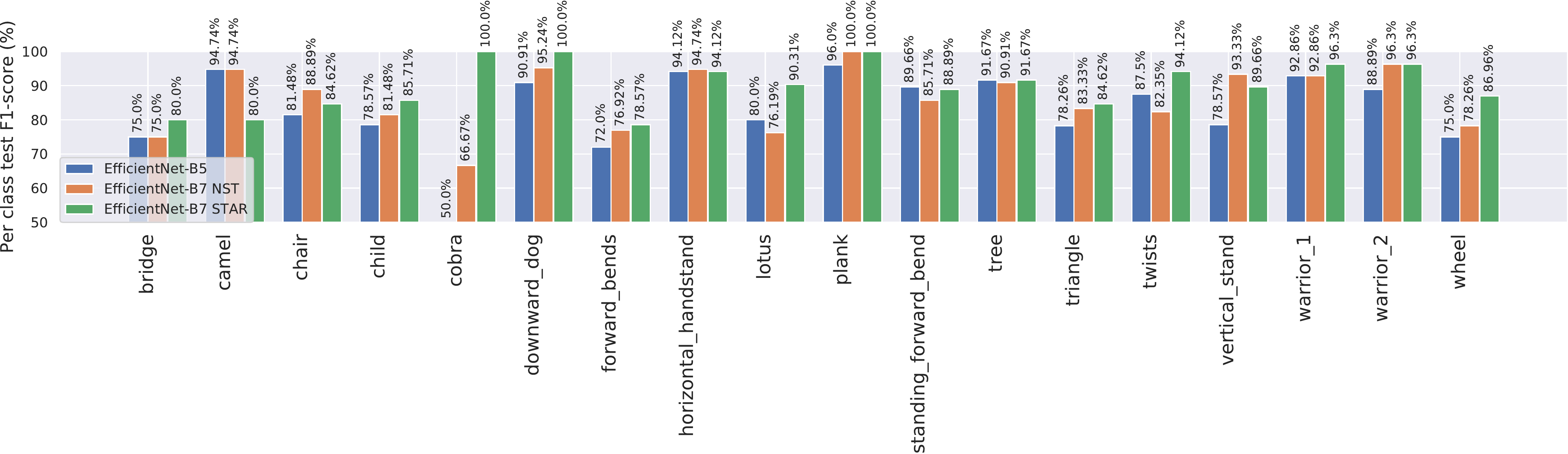}
\caption{Comparison of validation(top) and test (bottom) classification performance on the Yoga-Pose dataset for each class. EfficientNet-B5 is trained on the available training data, while EfficientNet-B7 NST and STAR are trained on a combination of the available labeled training data and generated pseudo-labels data by EfficientNet-B5.}
\label{Fig:yogapose_per_class_results}
\end{figure*}

\medskip\noindent\textbf{Yoga-Pose classification results.}\quad The overall performance results on the Yoga Pose dataset are summarized in Table~\ref{Tab:yogapose}. In this setting, the teacher model is the EfficientNet-B5 architecture trained on the labeled data. Similar to the Sport-or-not dataset, we generate 120K pseudo-labels from the Unlb-120 dataset and train the EfficientNet-B7 architecture in a similar fashion. Notice that STAR yields significant performance improvements with almost a 5.88\% relative improvement on the test set over the teacher model. Hence, the student model is able to make better predictions on both the validation and test set samples for individual classes, resulting in a much improved performance on the test set. This better performance is further illustrated in Figure~\ref{Fig:yogapose_per_class_results}, which shows the F1 score for each class of the Yoga-pose dataset on both validation and test sets. Figure~\ref{Fig:yogapose_per_class_results} (top) shows the performance on the validation set, and we can see that EfficientNet-B7 STAR outperforms the teacher model by a large margin on classes such as \textit{camel}, \textit{chair}, \textit{standing forward bend}, \textit{warrior 1}, and \textit{wheel}. We only see a drop in performance for the \textit{child} and \textit{twists} classes. In Figure~\ref{Fig:yogapose_per_class_results} (bottom), the performance on the test set is shown. Notice that the largest improvement using EfficientNet-B7 STAR is seen for the \textit{cobra} class with a F1-score of 100\%, whereas EfficientNet-B5 yields a score of 50\%. For the majority of the classes, a significant improvement is obtained using our STAR method compared to the baselines.

\begin{table}[!htb]
	\caption{Comparison of classification performance on the Yoga-Pose classification task. We train EfficientNet-B5 on the labeled training data. NST stands for Noisy Student Training. EfficientNet-B7 NST and STAR are trained with a combination of pseudo-labeled data, generated by EfficientNet-B5 on the unlabeled dataset, and the labeled data. Boldface numbers indicate the best performance.}
	\label{Tab:yogapose}
	\setlength{\tabcolsep}{3.8pt}
	\centering
	\medskip
	\begin{tabular}{@{}lcccccccccccc@{}}
		\toprule
		& \multicolumn{2}{c}{Accuracy (\%)} \\
		\cmidrule(lr){2-3}
		Method & Validation & Test\\
		\midrule
		EfficientNet-B5~\cite{tan2019efficientnet} ~ & 89.35 & 84.58 \\
		EfficientNet-B7 NST~\cite{xie2020self} ~ & 92.12 & 87.56 \\
		EfficientNet-B7 STAR ~ & \textbf{93.06} & \textbf{89.55} \\
		\bottomrule
		\hline
	\end{tabular}
\end{table}

\medskip\noindent\textbf{Sport classification results.}\quad Table~\ref{Tab:sport} shows the comparison results with baselines on the Sport dataset. As can be seen, STAR outperforms both EfficientNet-B5 and EfficientNet-B7 NST baselines by relative improvements of 21.43\% and 10.81\%, respectively. Notice that the test accuracy of EfficientNet-B5 is low compared to the validation accuracy, due in large part to the fact that the test set is class-balanced and hence accuracy provides an unbiased representation of type I and type II errors. Also, notice the large gap between the accuracy values on the validation and test sets for the baselines. As for the proposed method the gap is much smaller, suggesting that there is less overfitting during the hyperparameter optimization process.

\begin{table}[!htb]
\caption{Comparison of classification performance on the Sport classification task. We train EfficientNet-B5 on the labeled training data. EfficientNet-B7 NST and STAR are trained with a combination of pseudo-labeled data, generated by EfficientNet-B5 on the unlabeled dataset, and the labeled data. Boldface numbers indicate the best performance.}
	\label{Tab:sport}
	\setlength{\tabcolsep}{3.8pt}
	\centering
	\medskip
	\begin{tabular}{@{}lcccccccccccc@{}}
		\toprule
		& \multicolumn{2}{c}{Accuracy (\%)} \\
		\cmidrule(lr){2-3}
		Model & Validation & Test\\
		\midrule
		EfficientNet-B5~\cite{tan2019efficientnet} ~ & 83.51 & 66.25 \\
		EfficientNet-B7 NST~\cite{xie2020self} ~ & 85.44 & 72.60 \\
		EfficientNet-B7 STAR ~ & \textbf{89.28} & \textbf{80.45} \\
		\bottomrule
		\hline
	\end{tabular}
\end{table}

\subsection{Adversarial Robustness}
\begin{figure}[!htb]
\setlength{\tabcolsep}{.1em}
\centering
\begin{tabular}{cc}
\includegraphics[scale=.29]{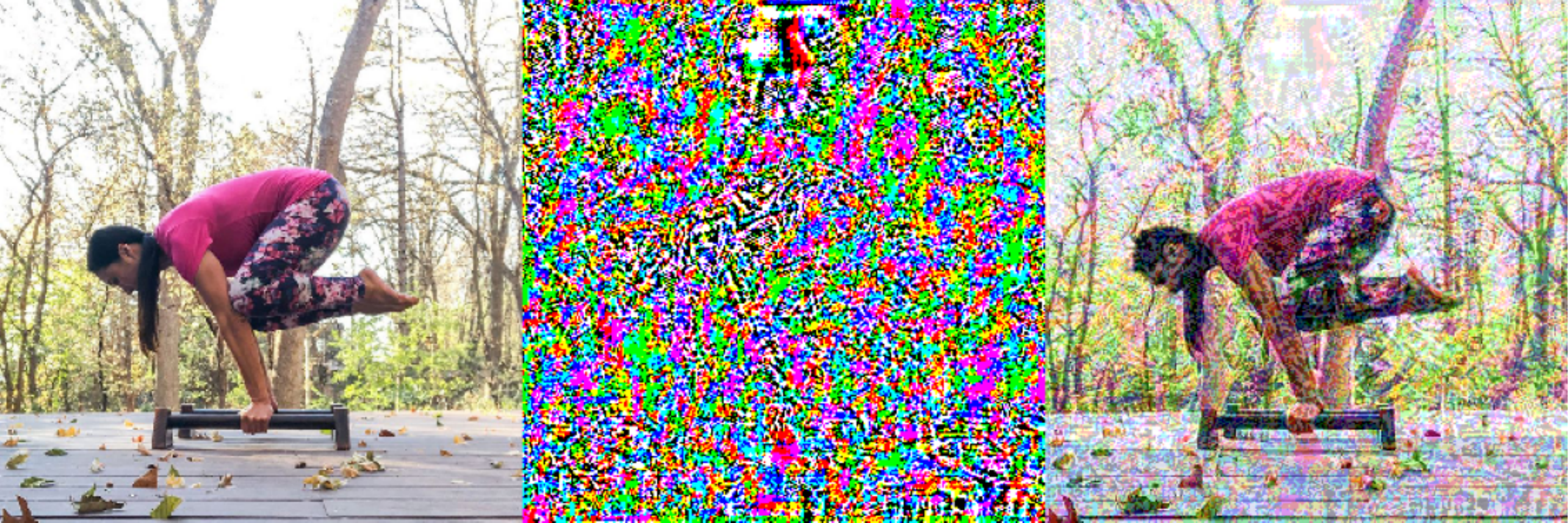}
\end{tabular}
\caption{Input image (left), perturbed image using FGSM attack (center), and resulting adversarial image (right). An $\epsilon = 0.13$ is used to ensure perturbations are small.}
\label{fig:adversarial}
\end{figure}

We also study the model performance against adversarial attacks. For each classification task (i.e. Sport-or-not, Yoga-Pose and Sport classification), we evaluate our best models both with and without NST or STAR against the fast gradient sign method (FGSM) attack~\cite{43405}, where the goal is to ensure misclassification. FGSM is a white box attack since it has complete access to the model being attacked. For a given input image, the method uses the gradient of the loss function with respect to the input image to create a new image that maximizes the loss, with the update on each pixel set to $\epsilon$. The new image is called the adversarial image. We show an illustration of an FGSM attach in Figure~\ref{fig:adversarial}, where the value of $\epsilon$ is set to 0.13.

\begin{figure}[!htb]
\setlength{\tabcolsep}{.1em}
\centering
\begin{tabular}{cc}
\includegraphics[scale=.36]{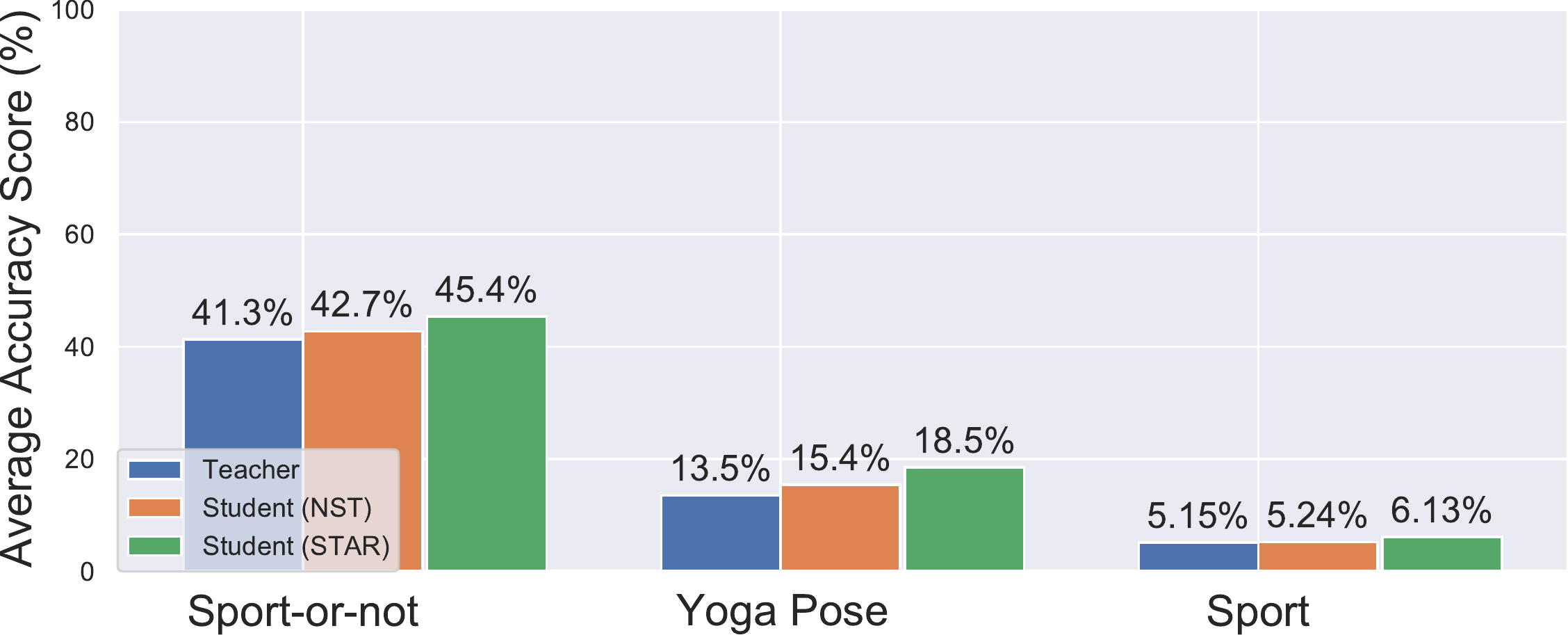}
\end{tabular}
\caption{Adversarial robustness results with $\epsilon = 0.13$. STAR improves adversarial robustness against white box FGSM attack even though we do not optimize the model for adversarial robustness. }
\label{fig:adv_res}
\end{figure}

In Figure~\ref{fig:adv_res}, we report the adversarial robustness results. As can be seen, our STAR method yields to significant improvements in terms of average accuracy despite the fact that no model optimization is performed for adversarial robustness. On the Yoga-Pose datset, we observe the highest improvement of 3.1 percentage points over NST and 5\% percentage points over the teacher model. For the Sport-or-not and Sport classification tasks, we see improvements of 2.7\% and 0.89\% over NST, and 4.1\% and 0.98\% over the teacher model.

\subsection{Ablation Study}
In this section, we study the importance of transfer learning in terms of runtime and memory, labeled data, and unlabeled data.

\medskip\noindent\textbf{Importance of transfer learning in self-training.} \quad Table~\ref{Tab:abl1} shows the importance of transfer learning in self-training. As can be seen, training student models using transfer learning not only yields better accuracy, but also reduces training time and computation overhead by significantly reducing the number of learnable parameters. Notice that using transfer learning in self-training requires 6x less training time over the baseline. The baseline method here is similar to NST, where the networks are trained from scratch. We believe that our approach is more practical for the industrial perspective in the sense that we can quickly develop models that can leverage unlabeled data in a semi-supervised fashion.

\begin{table}[!htb]
\caption{Ablation study of transfer learning on the Sport-or-not dataset. For the two cases, we train the EfficientNet-B5 using the pseudo-labels from Unlb-700 and 14K labeled images with and without ImageNet pretraining. TL stands for transfer learning and size refers to the number of learnable parameters. Boldface numbers indicate the best performance}
	\label{Tab:abl1}
	\setlength{\tabcolsep}{3.8pt}
	\centering
	\medskip
	\begin{tabular}{@{}lrrr@{}}
		\toprule
		%& \multicolumn{2}{c}{Accuracy (\%)} \\
		%\cmidrule(lr){2-3}
		Method & Seconds per Epoch & Size \\
		\midrule
		EfficientNet-B5 without TL~\cite{xie2020self} ~ & 250 & 30.6M\\
		EfficientNet-B5 with TL (Ours) ~ & \textbf{40} & \textbf{5.5M} \\
		\bottomrule
		\hline
	\end{tabular}
\end{table}

\medskip\noindent\textbf{Effect of varying labeled data in self-training.}\quad In order to understand the amount of required labeled data to achieve optimal performance, we vary the amount of labeled data available in the self-training process. As shown in Table~\ref{Tab:labeled}, the performance of the supervised EfficientNet-B5 model suffers a relative drop of 8.5\% when using only 1,000 samples for training. Interestingly, for EfficientNet-B7 STAR, even though there is a drop in performance, the classification performance is higher when using only 1,000 samples for training, where EfficientNet-B5 uses all available labeled data for training. This shows that in data scarcity scenarios, our STAR method can leverage unlabeled data to achieve optimal performance for the desired task.

\begin{table}[!htb]
\caption{Ablation study of labeled data for the Yoga-Pose classification task. We train the supervised classifier EfficientNet-B5 on the labeled data (1K, 2K, All), where All represents the entire training dataset. EfficientNet-B7 STAR is then trained on a combination of the labeled data and 120K pseudo-labeled data generated by EfficientNet-B5 from Unlb-120. Boldface numbers indicate the best performance.}
\label{Tab:labeled}
\setlength{\tabcolsep}{3.8pt}
\centering
\medskip
\begin{tabular}{@{}lcccccccccccc@{}}
\toprule
& \multicolumn{2}{c}{Test set accuracy (\%)} \\
\cmidrule(lr){2-3}
Labeled set size & EfficientNet-B5 & EfficientNet-B7 STAR\\
\midrule
1K ~ & 78.64 & 85.33 \\
2K ~ & 81.52 & 87.24 \\
All ~ & \textbf{84.58} & \textbf{89.55} \\
\bottomrule
\hline
\end{tabular}
\end{table}

\begin{figure}[!htb]
\setlength{\tabcolsep}{.05em}
\centering
\includegraphics[scale=.47]{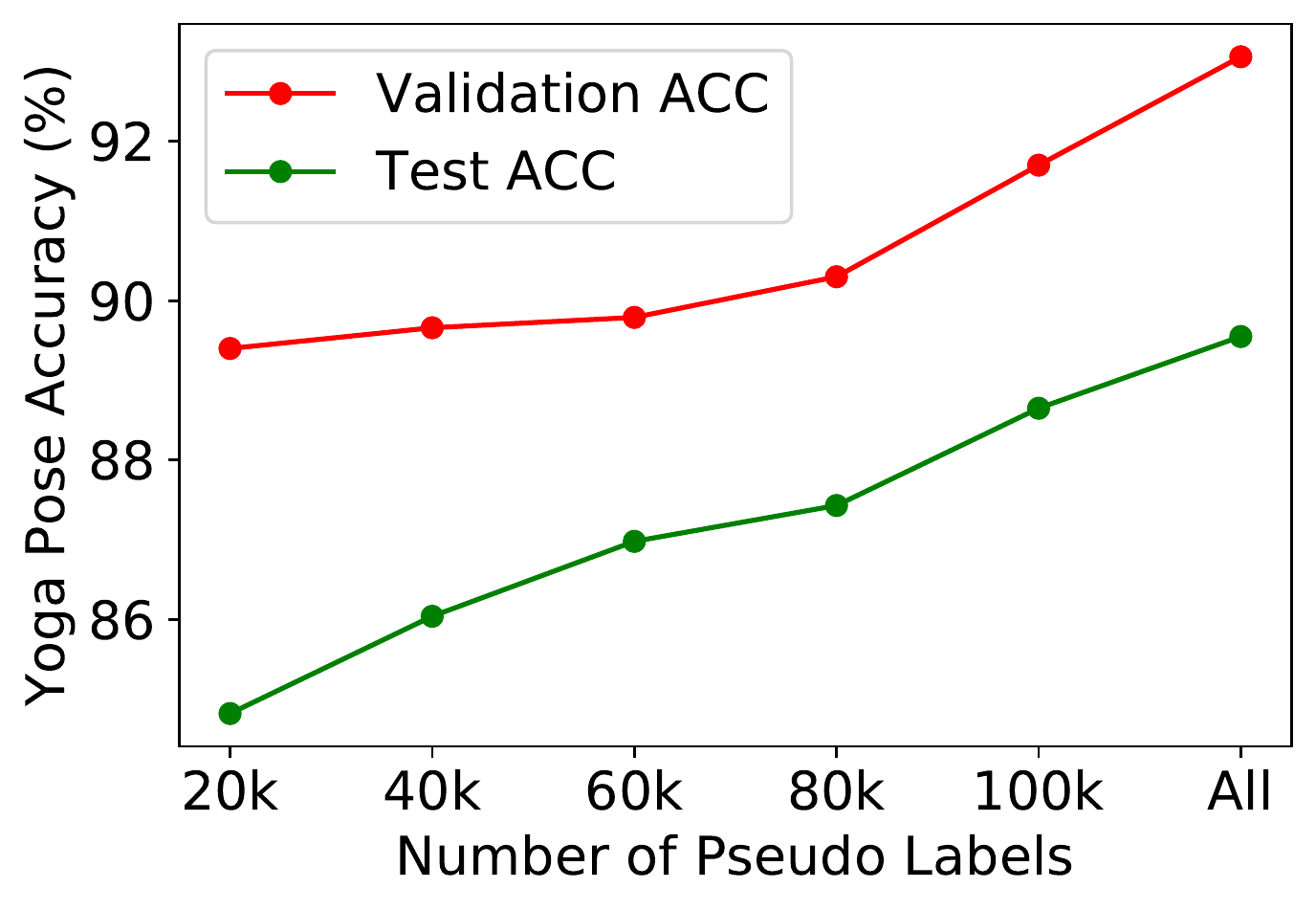} \\
\includegraphics[scale=.47]{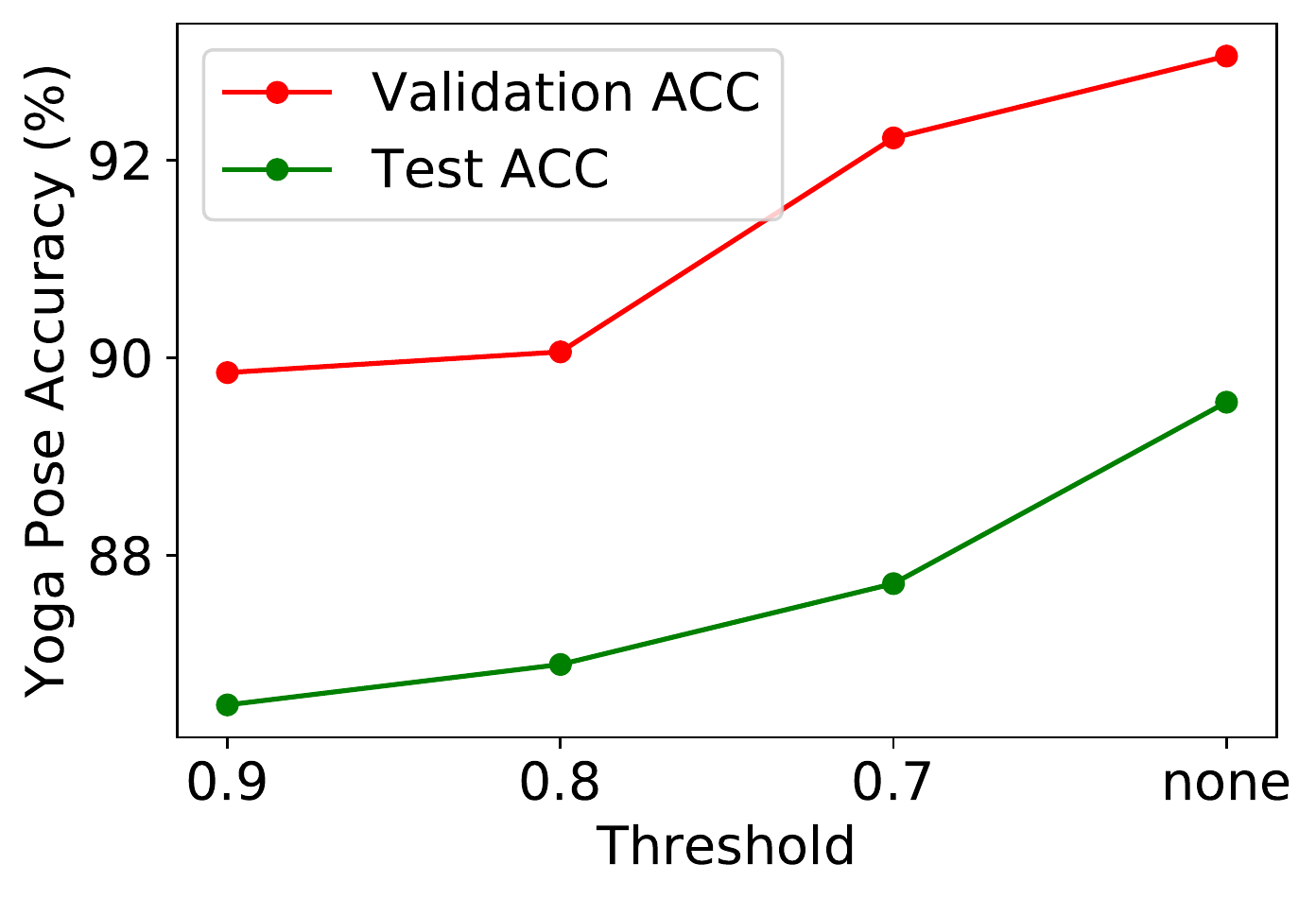}
\caption{Sensitivity analysis of EfficientNet-B7 STAR on the Yoga-Pose classification task by varying the number of pseudo-labels (top) and threshold (bottom). \textit{none} means that the lowest confidence predictions are taken in the pseudo-labeled data. In this case, all unlabeled images are part of the pseudo-labeled dataset.}
\label{fig:pseudo_label_and_threshold}
\end{figure}

\medskip\noindent\textbf{Effect of varying pseudo-labels in self-training.}\quad We study how the number of pseudo-labels generated from the unlabeled samples can affect classification performance in self-training. In Figure~\ref{fig:pseudo_label_and_threshold} (top), we can see that as the number of pseudo-labels grows, the classification performance shows an increasing trend in both validation and test sets. Moreover, adding more pseudo-labels would potentially further improve performance since the trend has not saturated yet.

In Figure~\ref{fig:pseudo_label_and_threshold} (bottom), we show how varying the threshold value affects performance. A threshold of 0.7 means that if the predicted label is below such a threshold, it is discarded from the pseudo-labeled dataset. Notice an upward trend when decreasing the threshold. This trend can be attributed to the fact that even though increasing the threshold may lead to higher quality pseudo-labeled data, more images are being discarded, resulting in a drop in performance. Moreover, we found using hard pseudo-labels (i.e. \textit{none}) on the unlabeled data yields better results compared to using soft pseudo-labels. This is also in line with findings in~\cite{xie2020self}, where soft pseudo-labels are shown to work better for out-of-domain unlabeled images.

\begin{figure}[!htb]
\setlength{\tabcolsep}{.1em}
\centering
\begin{tabular}{cc}
\includegraphics[scale=.54]{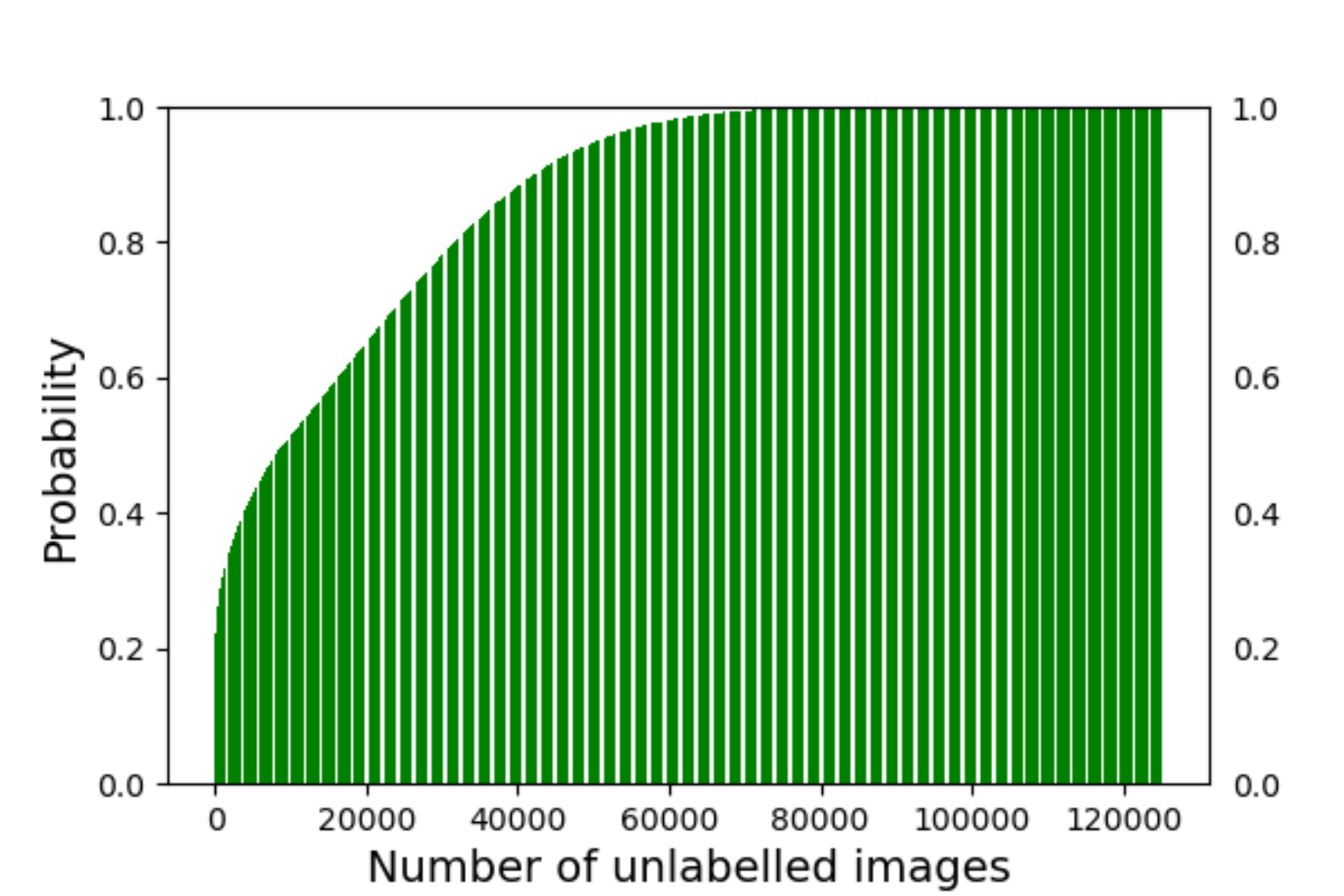}
\end{tabular}
\caption{Confidence scores for the teacher model on the Unlb-120 dataset for Yoga-Pose classification task. For the majority of the images, we obtain high confidence scores, indicating that images are relevant to this task and not out-of-domain.}
\label{fig:preds}
\end{figure}

In Figure~\ref{fig:preds}, we plot the raw probability scores by the teacher model on the unlabeled images used for the Yoga-Pose classification task. It is important to note that the plot shows that most of the pseudo-labels have high confidence due to a good teacher model. This is also attributed to the fact that the unlabeled images are acquired using search keywords or labels, which we ignored and treated them as unlabeled data. Hence, these images are somewhat relevant and not out-of-domain for the given task, and leaving relevant images out would degrade performance, as shown in Figure~\ref{fig:pseudo_label_and_threshold}.

\section{Conclusion} \label{Conclusion}
In this paper, we introduced STAR, a novel semi-supervised learning method that combines transfer learning and self-training with noisy student. The proposed framework is well suited for leveraging unlabeled images on a scale of thousands, and is efficient in terms of both runtime and memory, without compromising on accuracy. Extensive experiments show that STAR outperforms state-of-the-art methods, especially for multi-class classification tasks. Ablation studies also showed that leveraging transfer learning in the training process not only improves performance in terms of accuracy, but also requires 6x less compute time and 5x less memory. Moreover, we showed that STAR boosts robustness in visual classification models without specifically optimizing for adversarial robustness. While our focus in this work was on binary and multi-class classification use-cases, the proposed STAR method can be easily extended to other downstream tasks such as segmentation and object detection, which we intent to explore as future work.


\bibliographystyle{ACM-Reference-Format}
\bibliography{references}

%\appendix
%Lorem ipsum dolor sit amet, consectetur adipiscing elit. Morbi
%malesuada, quam in pulvinar varius, metus nunc fermentum urna, id
%sollicitudin purus odio sit amet enim. Aliquam ullamcorper eu ipsum
%vel mollis. Curabitur quis dictum nisl. Phasellus vel semper risus, et
%lacinia dolor. Integer ultricies commodo sem nec semper.
%Etiam commodo feugiat nisl pulvinar pellentesque. Etiam auctor sodales
%ligula, non varius nibh pulvinar semper. Suspendisse nec lectus non
%ipsum convallis congue hendrerit vitae sapien. Donec at laoreet
%eros. Vivamus non purus placerat, scelerisque diam eu, cursus
%ante. Etiam aliquam tortor auctor efficitur mattis.
%Nam id fermentum dui. Suspendisse sagittis tortor a nulla mollis, in
%pulvinar ex pretium. Sed interdum orci quis metus euismod, et sagittis
%enim maximus. Vestibulum gravida massa ut felis suscipit
%congue. Quisque mattis elit a risus ultrices commodo venenatis eget
%dui. Etiam sagittis eleifend elementum.
%
%Nam interdum magna at lectus dignissim, ac dignissim lorem
%rhoncus. Maecenas eu arcu ac neque placerat aliquam. Nunc pulvinar
%massa et mattis lacinia.

\end{document}